\pdfoutput=1

\documentclass[10pt,twocolumn,letterpaper]{article}
\PassOptionsToPackage{table,dvipsnames}{xcolor}
\usepackage{makecell}
\usepackage{multirow}
\usepackage{placeins}
\usepackage{pifont}
\usepackage{tcolorbox}
\usepackage{listings}
\usepackage{caption}

\usepackage{iccv}      

\definecolor{iccvblue}{rgb}{0.21,0.49,0.74}
\usepackage[pagebackref,breaklinks,colorlinks,allcolors=iccvblue]{hyperref}

\def\paperID{} 
\def\confName{}
\def\confYear{}

\title{VividListener: Expressive and Controllable Listener Dynamics Modeling for Multi-Modal Responsive Interaction}

\author{
Shiying Li$^{1*}$ \quad 
Xingqun Qi$^{2*\dagger}$ \quad 
Bingkun Yang$^1$ \quad 
Chen Weile$^1$ \quad 
Zezhao Tian$^1$ \quad \\
Muyi Sun$^{1\ddagger}$ \quad
Qifeng Liu$^2$ \quad 
Man Zhang$^1$ \quad 
Zhenan Sun$^3$ \\
\\
$^1$Beijing University of Posts and Telecommunications, Beijing, China \\
$^2$Hong Kong University of Science and Technology, Hong Kong, China \\
$^3$Institute of Automation, Chinese Academy of Sciences, Beijing, China 
}

\begin{document}
\twocolumn[{
    \maketitle
    \vspace{-3em}
    \begin{center}
      \includegraphics[width=\textwidth]{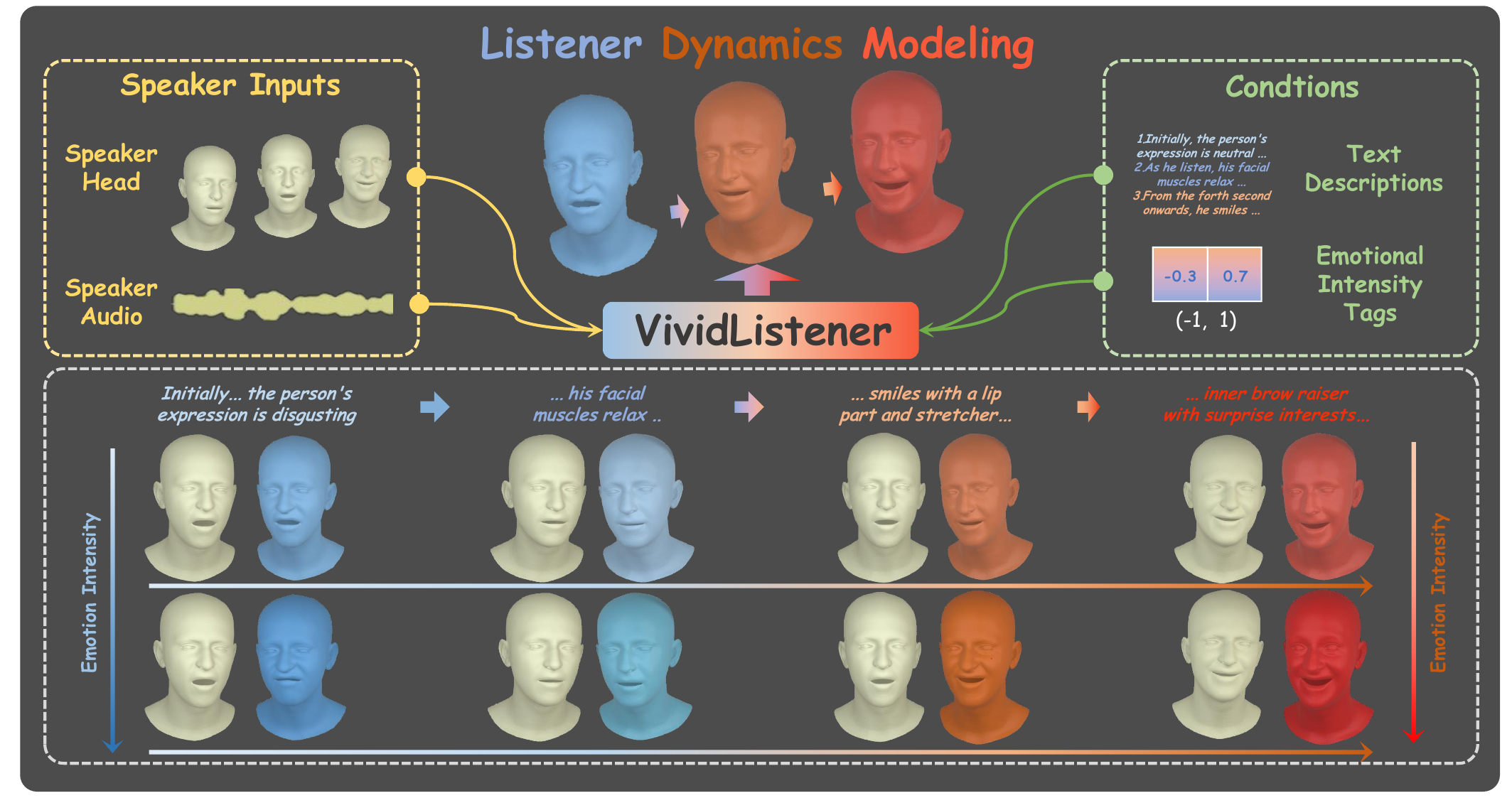}
      \vspace{-1em}
      \captionof{figure}{VividListener: listener dynamics modeling framework for multi-modal responsive interaction.  
      This framework inputs speaker head motions, speaker audios, the conditions of listener textual descriptions, and the emotional intensity tags, which outputs listener head dynamics sequences.
      The generated listener head could change with the expression descriptions (from light-blue/disgusting to red/excited), and exhibit varying degrees of facial emotions (from light to dark).}
      \label{fig:teaser}
    \end{center}
    \vspace{0em}
}]
\renewcommand{\thefootnote}{\fnsymbol{footnote}} 
\footnotetext[1]{Equal contribution.}
\footnotetext[2]{Project leader.}
\footnotetext[3]{Corresponding author.}

\begin{abstract}

Generating responsive listener head dynamics with nuanced emotions and expressive reactions is crucial for practical dialogue modeling in various virtual avatar animations. 
Previous studies mainly focus on the direct short-term production of listener behavior. 
They overlook the fine-grained control over motion variations and emotional intensity, especially in long-sequence modeling.
Moreover, the lack of long-term and large-scale paired speaker-listener corpora including head dynamics and fine-grained multi-modality annotations (\eg, text-based expression descriptions, emotional intensity) also limits the application of dialogue modeling.
Therefore, we first newly collect a large-scale multi-turn dataset of 3D dyadic conversation containing more than 1.4M valid frames for multi-modal responsive interaction, dubbed \textbf{ListenerX}.
Additionally, we propose \textbf{VividListener}, a novel framework enabling fine-grained, expressive and controllable \textbf{listener dynamics modeling}.
This framework leverages multi-modal conditions as guiding principles for fostering coherent interactions between speakers and listeners.
Specifically, we design the \textbf{Responsive Interaction Module} (RIM) to adaptively represent the multi-modal interactive embeddings.
RIM ensures the listener dynamics achieve fine-grained semantic coordination with textual descriptions and adjustments, while preserving expressive reaction with speaker behavior.
Meanwhile, we design the \textbf{Emotional Intensity Tags} (EIT) for emotion intensity editing with multi-modal information integration, applying to both text descriptions and listener motion amplitude.
Extensive experiments conducted on our newly collected ListenerX dataset demonstrate that VividListener achieves state-of-the-art performance, realizing expressive and controllable listener dynamics.

\end{abstract}

\section{Introduction}
Listener dynamics modeling aims to generate expressive and emotional head movements which respond to the speech content of the corresponding speakers. 
These non-verbal interactive behaviors~\cite{volonte2020effects} like facial expressions, nodding and blinking significantly facilitate message delivery during human daily communications. 
Meanwhile, modeling listener dynamics displays a wide range of applications in human-machine interaction~\cite{qi2024cocogesture, qi2024weakly}, embodied AI~\cite{duan2022survey}, robotics~\cite{soori2023artificial}, and virtual avatar animations~\cite{genay2021being}.

Previous studies on dialogue modeling have mostly focused on speaker generation (talking face/ head generation) and achieved satisfactory performances~\cite{chen2021usability,pei2024deepfake}. 
However, these works all concentrate on the individual speaker modeling, without considering the application of interactive scenarios, especially in modeling the response of listener.
Recently, several researchers have explored latent space modeling to generate listener motions from speaker information ~\cite{ng2022learning,zhou2022responsive,liu2023mfr,song2023emotional,liu2024customlistener,tran2024dim}. 
Some of them have employed the simplistic emotion labels to control the generation of listener motions and achieved good results~\cite {liu2023mfr,song2023emotional}. 
However, these approaches are typically limited to short-term head motion modeling and analysis, where the semantic and expression information are significantly insufficient and incomplete. 
Furthermore, the coarse perception of listener emotions with several basic categories exacerbates these shortcomings, which results in the lack of diversity and exhibits monotonous head expressions.
Thus, these methods have overlooked the fine-grained motion variants and emotion intensity within long dialogue sequences, which are more practical in real-world scenes. 
For example, the listener emotion may fluctuate dynamically during communication, shifting from excitement to frustration. 
Therefore, in this work, we introduce the new task of \textbf{Expressive and Controllable Listener Dynamics Modeling} for multi-modal responsive interaction in long conversational sequences, as illustrated in Fig.~\ref{fig:teaser}.

There are two main challenges in the listener dynamics modeling task:
1) Datasets containing fine-grained annotations of listener head movements and emotion intensity corresponding to speaker corpus are scarce. 
2) Modeling complex and variable emotion shifting of listener behavior is difficult, especially in long dialogue sequences from in-the-wild scenes.

To address the data scarcity problem, we first construct a new dataset of 3D head dyadic conversation which contains 1.4M frames of emotional speaker-listener head movements, dubbed \textbf{ListenerX}.
Specifically, we extract long-term and continuous dialogue segments from in-the-wild videos, ensuring that both the conversational participant heads remain fully visible throughout the interaction.
These segments should also include scenarios where the speaker articulates at least two clear sentences, while the listener reacts dynamically rather than monotonous.
To process these segments, we employ an advanced 3D facial avatar estimation method~\cite{danvevcek2022emoca} to extract high-quality facial expressions and head poses(FLAME~\cite{li2017learning}) for both conversational partners. 
Meanwhile, to support our insights into multimodal cue-guided modeling, we combine traditional detection models~\cite{chang2024libreface} with vision-language models~\cite{liu2024visual}to generate fine-grained and emotion-aware facial expression descriptions for continuous head frames. 
To control emotional intensity, we use facial emotion analysis techniques~\cite{toisoul2021estimation} to obtain Valence (how positive the emotion is) and Arousal (how calming or exciting it is), providing a continuous representation linked to emotional categories.
In this fashion, our ListenerX dataset covers high-quality and long-term listener head movements with corresponding fine-grained expression descriptions and emotion tags, which pave the way for various downstream tasks like human-human interaction analysis~\cite{stergiou2019analyzing, qi2024emotiongesture}, talking face generation~\cite{pei2024deepfake} and responsive interaction modeling~\cite{chen2021usability}.

Based on the \textbf{ListenerX} dataset, we propose \textbf{VividListener}, a novel framework that enables expressive and controllable listener head dynamics modeling with Diffusion Transformer (DiT).
Our key insight is to build coherent interaction between the speaker and responsive listener for listener modeling. 
Hence, our framework adaptively incorporates the multi-modal conditions as guiding principles via a diffusion-based generative model.
In particular, we propose the Responsive Interaction Module (RIM) to ensure temporal alignment between the generated listener movements and the speaker audio rhythms, while presenting fine-grained coordination \wrt text guidance. 
Here, we first model the joint interactive embeddings of the listener emotion descriptions and the integrated conversational cues provided by the given speaker audio and motion prior. 
In this manner, the discrete text-based descriptions are adaptively transformed into continuous temporal representations. 
Meanwhile, the speaker prior are seamlessly integrated into the learned joint embeddings. 
Moreover, we present Emotional Intensity Tags (EIT), a delicate design to achieve fine-grained emotion intensity control for listener dynamics. 
EIT together with text descriptions (listener motion information) serves as the input to learn a dynamic intensity tag through an integrated learning process.
Then, we can leverage the intensity tag as a controllable modulation mechanism to balance the dependency of generated facial dynamics.
The modulated features are finally fed into the DiT for producing expressive and controllable results.  

Overall, our contributions are summarized as follows:
\begin{itemize}[leftmargin=*]
    \item We introduce a new task of fine-grained controllable listener head dynamics modeling, cooperating with one newly collected large-scale 3D head dyadic conversation dataset, namely ListenerX. 
    \item We propose a novel framework, VividListener, that leverages multi-modal conditions as guiding principles to generate coherent and expressive listener head motions.
    \item We propose a Responsive Interaction Module to ensure coherent seamless integration of multi-modal conditions and a set of Emotional Intensity Tags to flexibly adjust emotional intensity, encouraging high-fidelity listener motion synthesis with desirable properties. 
    \item Extensive experiments demonstrate that our framework achieves superior performance against various competitors, displaying expressive and interactive coherent listener head dynamics. 
\end{itemize}

\begin{table*}[t]
\centering
\caption{Statistical comparison of our \textbf{ListenerX} against various counterparts. The "Long-term" means continuous and long-duration dialogue clips extracted from in-the-wild videos. The "Coarse" means simple emotional category labels. The abbreviation "Tag." stands for emotional intensity tags. The abbreviations "Sou." and "Rep" refer to the data source and the representation method. The scale represents the number of valid data frames in the dataset, without comparing the duration of the original videos.} 
\label{tab:dataset}
\footnotesize
\setlength{\tabcolsep}{2.5mm}{
\begin{tabular}{lcccccccccc}
\toprule
\multirow{2}{*}{\centering \textbf{Dataset}} & \multirow{2}{*}{\centering \textbf{Scale}} & \multirow{2}{*}{\textbf{Long-term}} & \multicolumn{4}{c}{\textbf{Modality}} & \multicolumn{2}{c}{\textbf{Annotation}}  & \multicolumn{2}{c}{\textbf{Acquisition}}   \\ \cmidrule(r){4-7}  \cmidrule(r){8-9} \cmidrule(r){10-11}
                        & &  &\textbf{Motion} & \textbf{Audio} & \textbf{Text} & \textbf{Tag.}  & \textbf{Coarse} & \textbf{Fine-grained}       & \textbf{Sou.}                     & \textbf{Rep.}   \\ \midrule \midrule
L2L Dataset~\cite{ng2022learning}\textcolor[HTML]{C0C0C0}{$_{CVPR}$}   & 0.75M   & \ding{55}(2S)                 & \ding{51}      & \ding{51}     & \ding{55}    & \ding{55}    & \ding{55}        & \ding{55}                  & YouTube & FLAME  \\
VICO~\cite{zhou2022responsive}\textcolor[HTML]{C0C0C0}{$_{ECCV}$}  & 0.1M   &\ding{55}(2S)                & \ding{51}      & \ding{51}     & \ding{55}    & \ding{55}    & \ding{51}        & \ding{55}                     &                  YouTube        & 3DMM   \\
Realtalk~\cite{geng2023affective}\textcolor[HTML]{C0C0C0}{$_{Arxiv}$}   & -  &\ding{55}(2S)             & \ding{51}      & \ding{51}     & \ding{55}    & \ding{55}    & \ding{55}        & \ding{55}                   &                        YouTube  & FLAME  \\
MDS Dataset~\cite{tamon2024listening}\textcolor[HTML]{C0C0C0}{$_{ICAI}$}   & 3M   & \ding{51}(6S-11S)                 & \ding{55}      & \ding{51}     & \ding{55}    & \ding{55}    & \ding{55}        & \ding{55}                  & Zoom & Video  \\  \midrule

\rowcolor[HTML]{ECF4FF} \textbf{ListenerX(Ours 2025)}  & \textbf{1.4M}      & \ding{51}(8S)        & \ding{51}      & \ding{51}     & \ding{51}    &\ding{51}    & \ding{51}        & \ding{51}                   &  \textbf{YouTube}                        & \textbf{FLAME}      \\ \bottomrule

\end{tabular}

}
\end{table*}

\section{Related Work}

\subsection{Listener Dynamics Modeling}

Listener dynamics modeling aims to generate expressive listener head movements in dyadic conversation. 
Early conversational avatar systems ~\cite{bohus2010challenges,bohus2010facilitating,sonlu2021conversational} mainly rely on rule-based methods to integrate facial expressions and gestures. 
Other studies ~\cite {huang2017dyadgan,nojavanasghari2018interactive} employ learning-based approaches to generate videos using 2D facial keypoints within a two-stage framework. 
Nevertheless, these methods exhibit limitations in adapting to dynamic environments and encounter considerable challenges related to scalability.

In recent years, advancements in 3D facial representation have significantly improved the ability to capture fine facial motions and have become the mainstream representation for listener motion generation.
Some early works focused on modeling the correlation between speaker information and listener motion~\cite{zhou2022responsive,ng2022learning,tran2024dim,ng2023can}. RLHG~\cite{zhou2022responsive} utilizes an LSTM-based autoregressive approach to generate simple listener reactions, such as nodding or smiling. L2L~\cite{ng2022learning}and DIM~\cite{tran2024dim}leverage VQ-VAE to learn discrete codebooks for listener motion. However, these models can only achieve coarse and uncontrollable listener motion generation.
Other works have achieved controllable generation based on input conditions~\cite{song2023emotional,liu2023mfr,liu2024customlistener}. ELP~\cite{song2023emotional} and MER-Net~\cite{liu2023mfr} enable listener motion generation using input emotions as control conditions, but the control over emotions is still limited to simple emotion labels.
CustomListener~\cite{liu2024customlistener} uses simple text input as a control condition which lacks fine-grained modeling of listener emotions. Meanwhile, all the aforementioned works focus on short-term dimensions, making it challenging to model the complex and dynamic listener reactions embedded with interactive and emotional information. 
In contrast, we introduce the long-sequence temporal modeling and integrate multi-modal annotations to achieve fine-grained emotional control in listener head dynamics generation.

\subsection{Multi-Modal Head Generation}
Multi-modal head generation aims to create 3D facial motion sequences based on input modalities such as speech audios, textual descriptions or emotional labels. 
Among these modalities, the audio modality has emerged as a primary focus of research due to the intrinsic correlation between speech and facial motion ~\cite{aneja2024facetalk,cudeiro2019capture,fan2022faceformer, richard2021meshtalk}.
For instance, Facetalk~\cite{aneja2024facetalk} and VOCA~\cite{cudeiro2019capture} focus on high-fidelity sequence modeling. However, their data acquisition processes are with high thresholds, which limits the practical applications. 
Other methods prioritize the accuracy of lip motion. 
For example, Meshtalk~\cite{richard2021meshtalk}and SD3DF~\cite{wu2023speech} disentangle the facial motion with the lip and non-lip regions.
However, these approaches struggle to coordinate overall facial emotions with lip movements, thereby overlooking the control of facial expressions.

Building on single-modality audio-driven approaches, several methods~\cite{peng2023emotalk,danvevcek2023emotional,sun2024diffposetalk, thambiraja2023imitator, zhao2024media2face}incorporate multimodal information to enable more precise control over 3D facial animation. 
EmoTalk~\cite{peng2023emotalk} and EMOTE~\cite{danvevcek2023emotional} introduce simple emotion labels to enhance the control of facial expressions.
DiffPoseTalk~\cite{sun2024diffposetalk} and Imitator~\cite{thambiraja2023imitator} enhance context awareness and generate stylistic motions from reference videos. Media2Face~\cite{zhao2024media2face} employs a diffusion-based method in the latent space guided by multi-modal inputs, including detailed text and images. 
However, most of these methods focus on the speaker modeling and ignore the interaction scenario, which overlook the interactive dynamics in conversational contexts.
And the conditional information extracted from multi-modal inputs sometimes remains coarse and discrete, neglecting fine-grained modeling. 
Inspired by the above, we introduce a multi-model, interactive and fine-grained listener head generation task, which enhances nuanced emotional interactions and enables emotional intensity control in dialogues.

\section{ListenerX Dataset}
\label{sec3}

To alleviate the scarcity of the 3D dyadic conversation dataset, we propose \textbf{ListenerX}, a newly collected large-scale multi-turn dataset for multi-modal responsive interaction, with the listener textual descriptions and emotional intensity tags. 

\subsection{Data Collection}
\label{sec3.1}
Considering the expensive and labor-consuming cost of 3D scanning or complex motion capture systems, similar to EMOTE\cite{danvevcek2022emoca}, we utilize in-the-wild videos as the data source and extract 3D representations by advanced facial avatar estimators. 
The original videos are sourced from interview shows and daily conversations, with each scenario accounting for half of the total frames, which we name InterviewX and DailyX.
InterviewX is characterized by structured contexts and stable emotional expressions, where listener responses primarily involve distinct non-verbal behaviors. 
In contrast, conversations in DailyX are more casual and emotionally dynamic, with listener facial movements exhibiting greater diversity and complexity.
Constructing a dataset tailored to our task emphasizes the acquisition of high-quality, long-duration, multi-turn interactive head movements between speakers and listeners, fine-grained textual annotations describing motion variations, and authoritative emotional intensity tags for facial expressions.

\subsection{Multi-Model Annotation Pipeline}
\label{sec3.2}

\noindent \textbf{3D Dyadic Conversation Reconstruction.} 
Firstly, we utilize face detection and voice source localization techniques~\cite{Chung16a} to produce long-term video segments in which the faces of both participants are simultaneously visible. 
In these segments, the roles of the speaker and listener remain consistent over continuous periods.
Then, we adopt the superior 3D facial estimator EMOCA~\cite{danvevcek2022emoca} to acquire the FLAME-based representation for each frame of both participants. 
Here, each clip in our dataset is unify cropped as \textbf{8 seconds} where the facial expressions and head pose movements are parameterized.
In this fashion, our dataset supports long-term multi-turn sequence modeling of interactive listener dynamics.

\noindent \textbf{Textual Descriptions.} 
Once we obtain the estimated motion clips, we annotate the varying facial expressions and emotions with fine-grained textual descriptions.
We observe that directly adopting the Vision-Language Models (VLM)~\cite{liu2024visual} on facial frames often produces inaccurate or fabricated descriptions, such as describing dimples that are not present in the faces. To this end, we incorporate the facial action unit detector~\cite{chang2024libreface} to extract high-intensity action units that serve as additional prompts for VLM. In this paradigm, we obtain accurate textual descriptions of the listener expressions.
Moreover, we conduct manual revisions to ensure the generated results are coherent and semantically aligned with facial sequences. 

\noindent \textbf{Emotional Intensity Tags.} 
To provide authority emotional intensity tags for expressive listener facial dynamics modeling, we incorporate facial affect analysis~\cite{toisoul2021estimation} in this step. 
Here, emotions are denoted as continuous dimensional representations regarding $V$ (Valence, how positive the emotion is) and $A$ (Arousal, how exciting the emotional display looks like)~\cite{russell1980circumplex}.
Compared to previous ones that directly apply discrete emotion classification, our approach offers a more reasonable reflection of human emotional expressions in interactive dialogues.

\subsection{Dataset Analysis}
\label{sec3.3}
Overall, our ListenerX contains a total of 6,683 videos with 1.4M frames for 3D conversational motion sequences. 
Each dialogue sequence includes the interactive 3D head movements of both participants, the speaker audio, fine-grained long-text emotion descriptions, and the emotion intensity tags of listener. 
Compared with the previous counterparts, ListenerX displays comprehensive modality attributes with fine-grained annotations as reported in Tab.~\ref{tab:dataset}. 
To the best of our knowledge, our ListenerX dataset is currently the largest for modeling long-term listener dynamics.

\begin{figure*}[t]
    \centering
    \includegraphics[width=\textwidth]{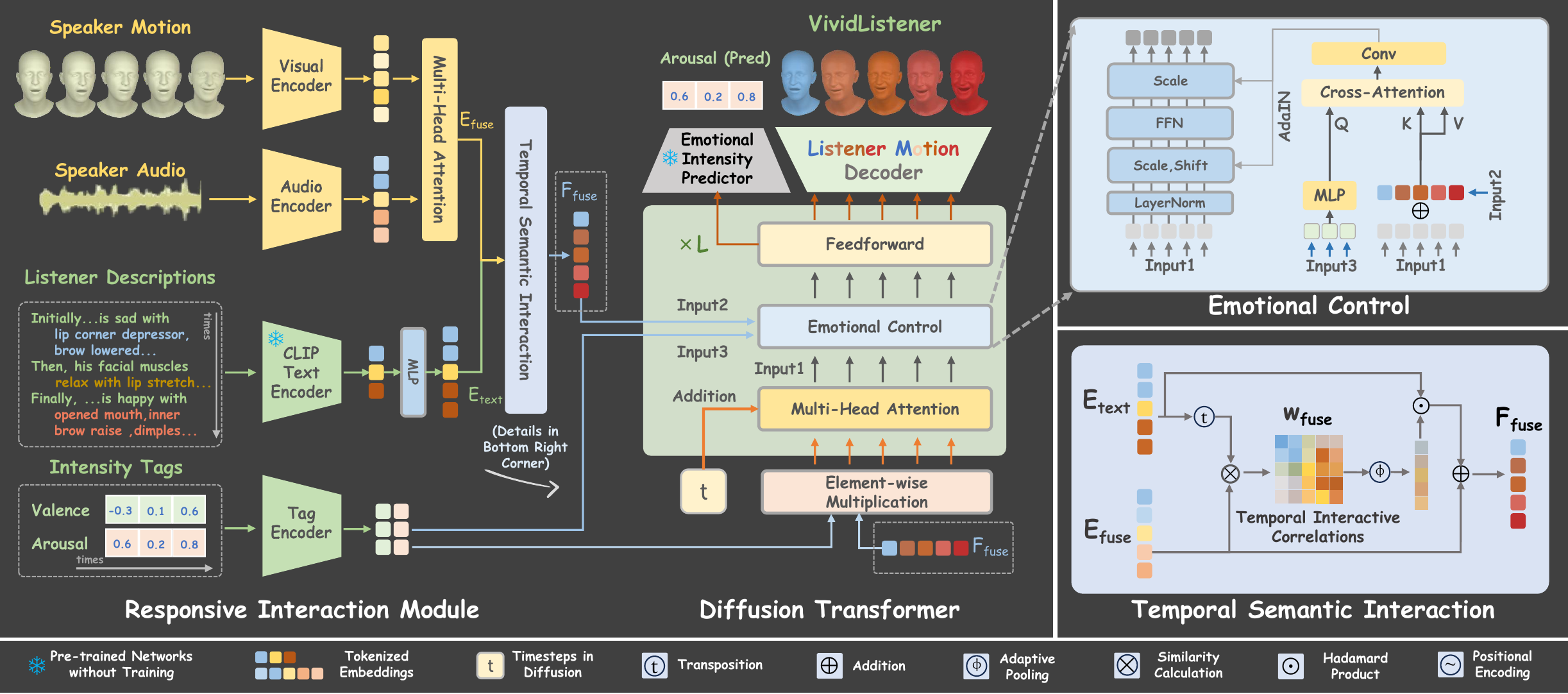}
    \caption{VividListener: listener dynamics modeling framework for multi-modal responsive interaction. This framework first integrates multi-modal inputs of speaker and listener through Responsive Interaction Module. Then, the fused features and conditions are fed into the DiT pipeline for the final listener dynamics modeling, which will change with the text descriptions and intensity tags, from cool(blue) tones to warm(red) tones. (Zoom in for better details.)}
    \label{fig:fig2}
\end{figure*}

\section{VividListener Framework}
\subsection{Problem Formulation}
Given the multi-modal inputs of the speaker motion \( S_m \), audio representation \( S_a \), fine-grained text descriptions $text$, and emotional intensity tags \( \sigma \), the goal of our framework is to generate expressive and controllable listener head motions $H$.
In practice, we leverage the parameterized FLAME model~\cite{li2017learning} to represent facial expressions and head poses.
We tackle this challenging problem with a specific-designed framework named VividListener.
The overall objective is expressed as
\begin{align}
H =\mathrm{VividListener}(S_m,S_a,text,\sigma)
\label{eq1}
\end{align}


\subsection{Listener Head Dynamics Modeling}
To generate high-fidelity and diverse results, our VividLinstener adopts the Diffusion Transformer (DiT)~\cite{peebles2023scalable} as the backbone, owing to its scalability and superior performance on sequence modeling, displayed in Fig.~\ref{fig:fig2}.
With the text descriptions of the listener movement as prompts, our framework first integrates the audio signals and head motion cues of the conversational speaker to produce fine-grained responsive guidance through the Responsive Interaction Module. 
Then, incorporated with the multi-modal responsive guidance, we employ the Emotional Intensity Tags to adaptively control the emotion intensity of generated results. 

\noindent \textbf{Responsive Interaction Module.}
To ensure the semantic coherent with text descriptions while preserving the vivid reaction, we propose Responsive Interaction Module (RIM) to model the temporal representations among multi-modal conditions. 
As illustrated in Fig.~\ref{fig:fig2}, the speaker features including motion and audio, are firstly fused via a bidirectional Multi-Head Attention (MHA) mechanism. 
Once we acquire the fused reactive features, we consider taking textual motion scripts as the semantic descriptions to further enhance the generation of corresponding listener movements. 
Here, we utilize a pre-trained CLIP ~\cite{radford2021learning} as the text encoder to obtain sequence-aware semantic embeddings. 
The aforementioned reactive features and semantic embeddings are then integrated into a joint representation via Temporal Semantic Interaction for modeling semantic associations and temporal dynamics. 

In particular, we model the temporal relevance between the speaker reactive features and semantic text embeddings via a joint similarity matrix, as shown in Fig.2 (bottom right corner).
We aim to exploit the temporal interactive correlations to boost the speaker motion cues for drawing listener synthesis. 
Once we attain this matrix, we conduct temporal-wise adaptive pooling to acquire a set of normalized (norm) learnable weight coefficients. 
These coefficients represent the activated blending weights aligned with the text embeddings. 
Then, we obtain the semantic-enhanced sequential speaker reaction features by:
\begin{align}
F_{\mathrm{fused}} & = E_{\mathrm{fused}} + E_{\mathrm{text}} \odot Norm(\text{AdaPool} \left ( W_\mathrm{fuse}\right  ))
\notag
\\
W_\mathrm{fuse} &= E_{\mathrm{fused}} \otimes {E_{\mathrm{text}}}^{'}
\label{eq2}
\end{align}
Where $W_\mathrm{fuse}\in\mathcal{R} ^{L\times L}$ is the temporal interactive correlation matrix, and $L$ is the sequence length. 
AdaPool indicates the adaptive maxpooling operation \cite{lecun1998gradient}.
$\odot$ means Hadamard product, $\otimes$ means matrix multiplication for similarity, and $'$ indicates the transposition operation. 
The semantic enhanced multi-modal features are further fed into our DiT-based denoiser. 
In this manner, the temporal interaction authority of the generated listener head is well-preserved.

\noindent  \textbf{Emotional Intensity Tags.}
For achieving fine-grained emotion intensity control in listener dynamics, the Emotional Intensity Tags serve as a key component in the VividListener framework. 
By integrating multi-modal information (\ie textual descriptions and motion features), EIT are injected into both the input conditions and intermediate layers of the framework. 
It provides emotional intensity control in a dynamically modulated manner.

Specifically, we first aggregate the emotional intensity tags with semantic-enhanced reactive features via element-wise multiplication to produce the emotion guidance fed into our framework.
Different from directly stacking the DiT blocks for listener dynamics generation, we further incorporate the aforementioned intensity tags and fused reactive motion cues into a specific-designed Emotional Control layer to draw a fine-grained facial effect.
As depicted in Fig.~\ref{fig:fig2} (upper right), the enhanced reactive fused features (\ie Input 2) are integrated with intermediate listener motion embeddings (\ie Input 1). 
Then we leverage the emotion tag representations as the query $Q$ to match the key features $K$ and value features $V$ via the cross-attention mechanism.
Along with this operation, we further obtain the updated emotional guidance embeddings by a convolution layer. 
Here, we take a consideration leveraging the emotional guidance as the modulated indicators which are exploited to boost the listener motion features via an adaptive instance normalization (AdaIN) layer~\cite{huang2017arbitrary}. By conducting this, we derive the listener dynamics features as:
\begin{align}
F_{listener} = \text{AdaIN}\left (F_{listener}, conv(F_{emo})\right ),
\label{eq4}
\end{align}
where $F_{listener}$ denotes listener motion features, and $F_{emo}$ indicates the emotion guidance produced by cross-attention.

\subsection{Objective Functions}
During the training phase, the entire framework is trained with an end-to-end manner. Given the diffusion timestep \(t\), the current speaker motion \(S_m\) and audio \(S_a\), the textual descriptions $text$, emotional intensity tags \(\sigma\), and the noised listener head motion $H^{(t)}$, the denoiser is designed to generate continuous listener head dynamics. The denoising process is constrained by a simple objective:
\begin{align}
\mathcal{L}_{\text{simple}} = \mathbb{E}_{H, t, \epsilon} \left[ \left\| H - \mathrm{VividL}(H^{(t)}, S_m, S_{a}, text,\sigma,t) \right\|_2^2 \right] ,
\label{eq6}
\end{align}
Where \(\epsilon \sim \mathcal{N}(0, I)\) is the added random Gaussian noise, \(H^{(t)} = H + \gamma_{(t)}\epsilon\) represents the gradual noise addition process at step \(t\). \(\gamma_{(t)} \in (0, 1)\) is a constant hyper-parameter.

\noindent \textbf{Emotional Intensity Predictor.} To ensure that the generated listener emotion displays coherent alignment with the input emotional intensity tags, we employ a pre-trained 3D emotional intensity predictor to produce the corresponding emotion tag for our results. 
The emotional intensity loss is defined as follows:
\begin{align}
\mathcal{L}_{\text{emotional}} = \|\sigma - \mathcal{P}_{\text{emotional}}(\hat{H})\|_2^2 ,
\label{eq6}
\end{align}
where $\hat{H}$ is our generated results. \(\mathcal{P}_{\text{emotional}}\) denotes emotion intensity predictor. 

Moreover, we utilize the velocity loss $\mathcal{L}_{vel}$ to provide supervision on the smoothness~\cite{zhang2022motiondiffuse}.
Finally, our overall objective function is defined as follows:
\begin{align}
\mathcal{L}_{\text{total}} = \lambda_{\text{simple}} \mathcal{L}_{\text{simple}} + \lambda_{\text{emotional}} \mathcal{L}_{\text{emotional}} + \lambda_{\text{vel}}\mathcal{L}_{\text{vel}}.
\label{eq7}
\end{align}
\(\lambda_{\text{simple}}\), \(\lambda_{\text{emotional}}\), and \(\lambda_{\text{vel}}\) are the trade-off coefficients.

\begin{figure}[t]
    \centering
    \includegraphics[width=0.48\textwidth]{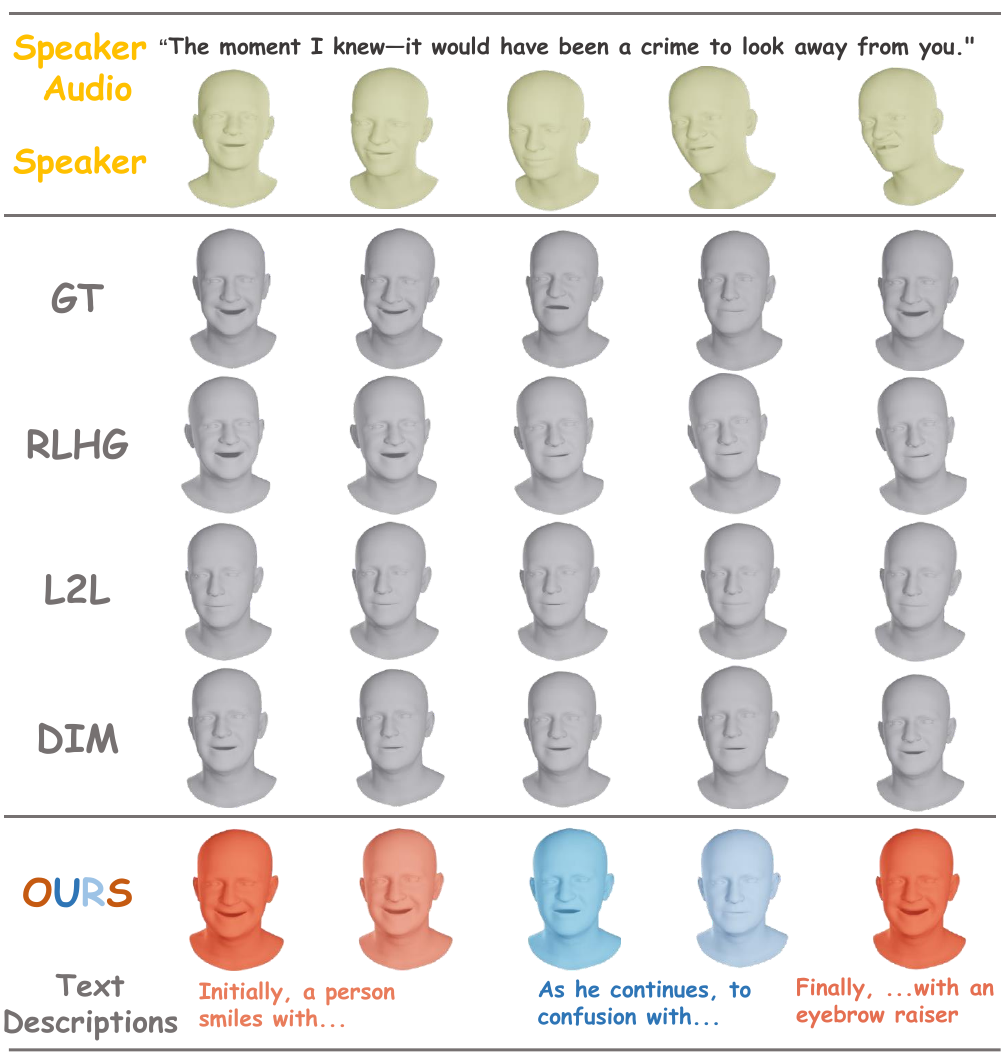}
    \caption{\textbf{Visual comparisons on ListenerX.} We present visualizations of listener motions generated on ListenerX compared to various state-of-the-art methods. Unlike other approaches, our VividListener input incorporates fine-grained textual descriptions (shown in the last row).}
     \label{fig:fig3}
\end{figure}
 \section{Experiments}

 \begin{figure*}[t]
    \centering
    \includegraphics[width=0.98\textwidth]{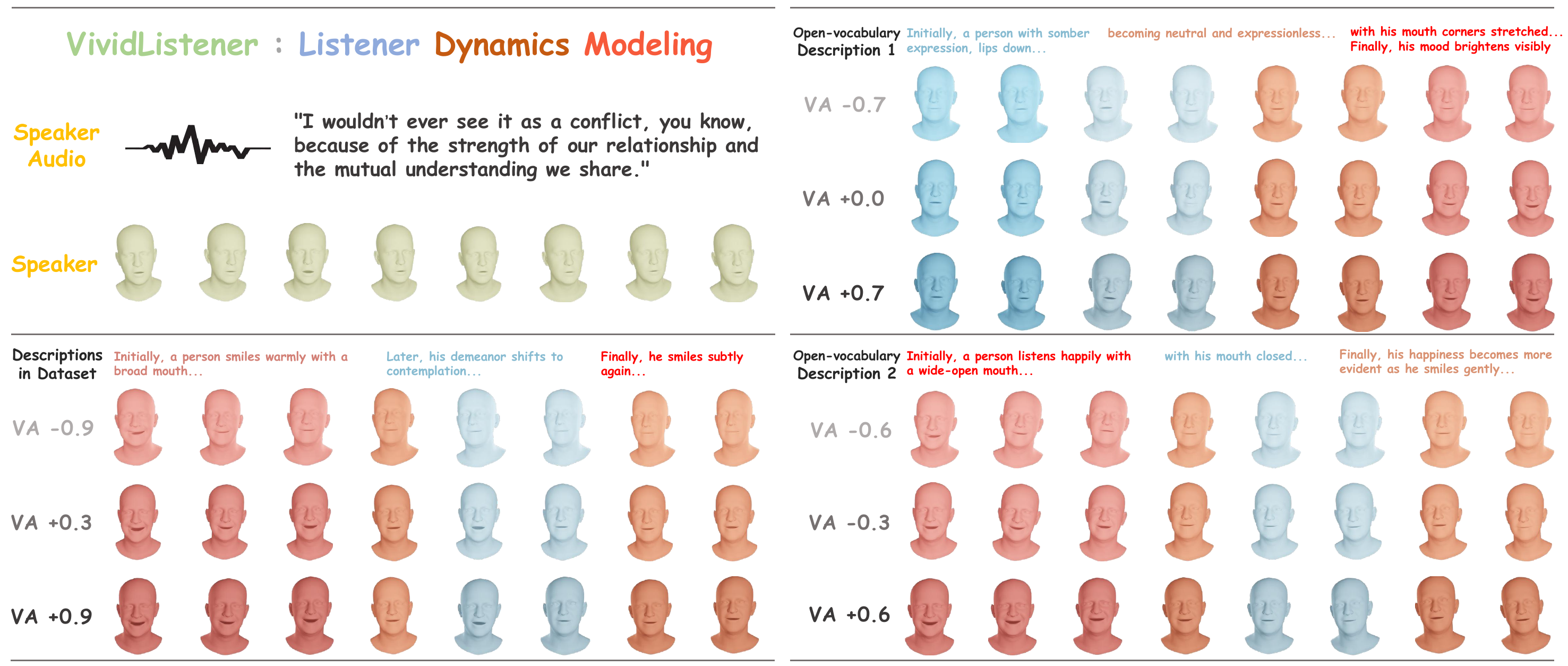}
    \caption{ \textbf{Conditional Control Results.} We generate listener head dynamics by inputting the speaker audio and head motion information (shown in the top-left corner) alongside dataset-based emotion descriptions (shown in the bottom-left corner). Additionally, we utilize open-vocabulary text inputs to further enrich the generation process. Each example incorporates three VA values (e.g., -0.9, +0.3, +0.9) to control emotional intensity levels (Please Zoom in for better details.)}
    \label{fig:fig4}
\end{figure*}
\subsection{Experimental Settings}
\noindent  \textbf{Implementation Details.}
In our experiments, we set the total generated sequence length N = 240 with the normalized fps=30. 
\(S_a\) is initially an audio waveform, converted into 128 × 480 mel-spectrograms using a 512 FFT window and a 160 hop length. 
The textual descriptions $text$ of the listener are encoded with the CLIP model~\cite{radford2021learning}. 
The listener emotional intensity tags \( \sigma \) are obtained at the frequency of 5 Hz.
In the training stage, we empirically set \(\lambda_{\text{simple}}\)=2, \(\lambda_{\text{emotional}}\)=0.2, \(\lambda_{\text{vel}}\)=0.8. The initial learning rate is set to \(1 \times 10^{-5}\) with AdamW optimizer~\cite{loshchilov2017decoupled}.

\noindent  \textbf{Evaluation Metrics.}
To comprehensively evaluate the realism, synchrony, and diversity of the generated listener dynamics, we introduce several metrics, including Fréchet Distance (FD), Paired Fréchet Distance (P-FD), Mean Squared Error (MSE), Shannon Index (SID), Variance (Var), and Residual Pearson Correlation Coefficient (rPCC). For a detailed description, please refer to the supplementary materials.

\begin{table*}[t]
\centering
\caption{
The top table presents a comparison with state-of-the-art methods on the ListenerX dataset, while the bottom table illustrates cross-scenario inference results.$\uparrow$ means the higher the better,  and $\downarrow$ indicates the lower the better.}
\label{tab:result}
\footnotesize
\setlength{\tabcolsep}{1.3mm}{
\begin{tabular}{llcccccccccccc}
\toprule
\multirow{2}{*}{Dataset Scenario}  & \multirow{2}{*}{Methods} & \multicolumn{2}{c}{FD$\downarrow$} & \multicolumn{2}{c}{P-FD$\downarrow$} & \multicolumn{2}{c}{MSE$\downarrow$} & \multicolumn{2}{c}{SID$\uparrow$} & \multicolumn{2}{c}{Var$\uparrow$} & \multicolumn{2}{c}{rPCC$\downarrow$}  \\  \cmidrule(r){3-4} \cmidrule(r){5-6} \cmidrule(r){7-8} \cmidrule(r){9-10} \cmidrule(r){11-12} \cmidrule(r){13-14}
                                   &                         & Exp & Pose              & Exp & Pose                & Exp & Pose               & Exp & Pose              & Exp & Pose              & Exp & Pose                 \\  \midrule \midrule
\multirow{4}{*}{ \makecell{InterviewX+DailyX\\(ListenerX)} }       
& RLHG~\cite{zhou2022responsive}                     & 5.842    & 0.021                  & 6.164    & 0.020                    & 0.133    & 0.003                   & 4.765     & 3.797                  & 0.212    & 0.015                 & 0.033     & 0.0052                     \\
& L2L~\cite{ng2022learning}                  &11.510     & 0.115                  & 11.770     & 0.113                     & 0.260     & 0.127                   & 1.930     &1.131                   &0.182     & 0.014                  & 0.028     & 0.134                     \\
& DIM~\cite{tran2024dim}                                      & 7.805    & 0.046                    & 8.291    & 0.042                   & 0.212     &0.006                   &4.050     & 3.441                  & 0.216    & 0.013   & 0.074    & 0.011                   \\
& \cellcolor[HTML]{ECF4FF}\textbf{Vividlistener (Ours)}    & \cellcolor[HTML]{ECF4FF} \textbf{3.792}    & \cellcolor[HTML]{ECF4FF}\textbf{ 0.036}                 & \cellcolor[HTML]{ECF4FF}\textbf {4.177}    & \cellcolor[HTML]{ECF4FF}\textbf{ 0.038 }              & \cellcolor[HTML]{ECF4FF}\textbf{ 0.120 }  & \cellcolor[HTML]{ECF4FF}  \textbf{0.006}           & \cellcolor[HTML]{ECF4FF}\textbf{ 4.803 }   & \cellcolor[HTML]{ECF4FF} \textbf{3.830}                 & \cellcolor[HTML]{ECF4FF}\textbf{ 0.210}    & \cellcolor[HTML]{ECF4FF}\textbf{ 0.015}                 & \cellcolor[HTML]{ECF4FF} \textbf{0.040}   & \cellcolor[HTML]{ECF4FF}   \textbf{0.037}                  \\ \midrule
\multirow{4}{*}{\makecell{DailyX(Train) \\ InterviewX(Test)}}            
& RLHG~\cite{zhou2022responsive}                     & 41.510   & 0.122                  &41.961     &0.126                     & 0.854    &0.028                    &1.040     &2.452                   & 0.882    & 0.004                  & 0.068    & 0.171                     \\
& L2L~\cite{ng2022learning}                     & 44.020    & 0.053                  &44.644     & 0.058                    &0.920     & 0.011                   & 1.452    & 2.630                  & 0.884    & 0.005                   & 0.043    &  0.154                    \\
& DIM~\cite{tran2024dim}                      &44.643     &0.142                   & 44.870    & 0.147                    & 0.913    & 0.020                   & 0.910    & 2.474     & 0.882    & 0.005       &0.043     & 0.138                     \\
& \cellcolor[HTML]{ECF4FF}\textbf{Vividlistener (Ours)}
& \cellcolor[HTML]{ECF4FF} \textbf{16.309}    & \cellcolor[HTML]{ECF4FF} \textbf{0.023}               
& \cellcolor[HTML]{ECF4FF} \textbf{16.890}    & \cellcolor[HTML]{ECF4FF}\textbf {0.028}                  
& \cellcolor[HTML]{ECF4FF}\textbf{0.363}    & \cellcolor[HTML]{ECF4FF} \textbf {0.003 }             
& \cellcolor[HTML]{ECF4FF}\textbf {1.460}  & \cellcolor[HTML]{ECF4FF} \textbf{1.506 }                
& \cellcolor[HTML]{ECF4FF} \textbf{0.821}   & \cellcolor[HTML]{ECF4FF} \textbf {0.005}              
& \cellcolor[HTML]{ECF4FF}\textbf{0.094}   & \cellcolor[HTML]{ECF4FF} \textbf  {0.168}               \\ \midrule
 
\end{tabular}
}
\end{table*}

\subsection{Quantitative Analysis}  

\noindent  \textbf{Comparisons with the SOTA counterparts.} 
To verify the effectiveness of our method, we compare the VividListener with several SOTA counterparts for listener modeling, L2L~\cite{ng2022learning}, RLHG~\cite{zhou2022responsive}, and DIM~\cite{tran2024dim}. 
These models are re-trained and tested on our dataset based on the source code released by the authors. 
The experiments are conducted in two groups.
We first apply the experiments upon the whole ListenerX dataset to verify the effectiveness of our framework on complex scenarios.
Then, to verify the model generalizability, DailyX is used for training, and InterviewX is used for testing. 
As shown in Tab.~\ref{tab:result}, VividListener consistently outperformed the competing models in both experimental settings. Notably, in the second experiment, our model achieved a significant reduction in FD by $25.21\%$ compared to the sub-optimal counterparts, demonstrating the superior generalization of our framework.

\noindent  \textbf{Ablation Study.} 
To further evaluate the effectiveness of our network design, we conduct the ablation studies on different components and input conditions as variations.
\textbf{For verifying the effects of the Vividlistener modules,} we conduct ablation studies on the Temporal Semantic Interaction (TSI), Emotional Control (EC), and the Emotional Identity Predictor (EIP). 
As shown in Tab.~\ref{tab:Ablation block}, removing TSI weakens semantic association and temporal modeling between text and speaker information, degrading global control and overall performance.
When EC is removed, SID and rPCC decrease, indicating its role in enhancing interactive information and control over listener motion.
Lastly, removing the EIP results in a decline in the Var metric, indicating its role in maintaining intensity consistency for higher-quality emotional expression.
\textbf{For verifying the effects of the conditional input guidances,} 
we conduct ablation studies by removing the text input and the emotion intensity tags, as show in Tab.~\ref{tab:Ablation condition}. Removing the text disrupts the semantic guidance provided by textual descriptions, leading to semantic alignment loss in the generated listener motion. 
This results in noticeable degradation in the rPCC and FD metrics, as the model fails to establish coherent interactions between speaker cues and listener responses. 
Similarly, removing the emotion intensity tag weakens the model ability to modulate the intensity of the generated motion, reducing the diversity and emotional expressiveness (SID and Var metrics) of the output. 
\begin{table}
\centering
\caption{Ablation study of Modules on ListenerX dataset. 
}
\label{tab:Ablation block}
\setlength{\tabcolsep}{3.2 mm}{%
\footnotesize
\begin{tabular}{lcccc}
\toprule
& \multicolumn{4}{c}{ListenerX}  \\ \cmidrule{2-5} 
\multirow{-2}{*}{Methods} & FD$\downarrow$ & SID$\uparrow$ & Var$\uparrow$ & rPCC$\downarrow$       \\ \midrule \midrule
w/o TSI                  & 6.38    & 1.82     & 0.18    & 0.09            \\
w/o EC          &4.00     & 4.72     & 0.18    & 0.05            \\
w/o EIP         & 5.95   & 2.07     & 0.09    & 0.08            \\
\rowcolor[HTML]{ECF4FF} \textbf{Vividlistener (Full)}      & \textbf{3.79}    & \textbf{4.80}     & \textbf{0.21}    & \textbf{0.04}           \\
\bottomrule
\end{tabular}%
}
\vspace{-2mm}
\end{table}

\begin{table}
\centering
\caption{Ablation study of input conditions on ListenerX dataset. 
}
\label{tab:Ablation condition}
\setlength{\tabcolsep}{3.2 mm}{%
\footnotesize
\begin{tabular}{lcccc}
\toprule
 & \multicolumn{4}{c}{ListenerX}  \\  \cmidrule{2-5} 
\multirow{-2}{*}{Input Conditions}                                 & FD$\downarrow$ & SID$\uparrow$ & Var$\uparrow$ & rPCC$\downarrow$       \\ \midrule \midrule
w/o Text                         & 4.83     & 4.14     & 0.18    & 0.05            \\
w/o Tag                          & 4.79     & 4.34     & 0.10    & 0.04            \\
\rowcolor[HTML]{ECF4FF} \textbf{Vividlistener (Full)}                   & \textbf{3.79}     & \textbf{4.80}     & \textbf{0.21}    & \textbf{0.04}            \\

\bottomrule
\end{tabular}%
}
\vspace{-4mm}
\end{table}


\subsection{Qualitative Analysis} 
\noindent  \textbf{Comparisons with the SOTA counterparts.}
To fully demonstrate the superior performance of our VividListener framework, we present visualized keyframes generated by our method compared with various counterparts on the ListenerX dataset. 
As shown in Fig.~\ref{fig:fig3} , our method produces more vivid and accurate listener head motions compared to other approaches. 
Specifically, we observe that L2L tends to generate rigid and monotonous results in long-sequence generation. 
RLHG achieves higher diversity, however, its facial expressions and head dynamics are limited in terms of expressiveness. 
DIM generates richer motions, yet the results lack synchronization with the speaker rhythm, leading to poor interactive dynamics. Please refer to the supplementary materials for the comparative results in cross-scenario inference.

\begin{figure}[t]
    \centering
    \includegraphics[width=0.48\textwidth]{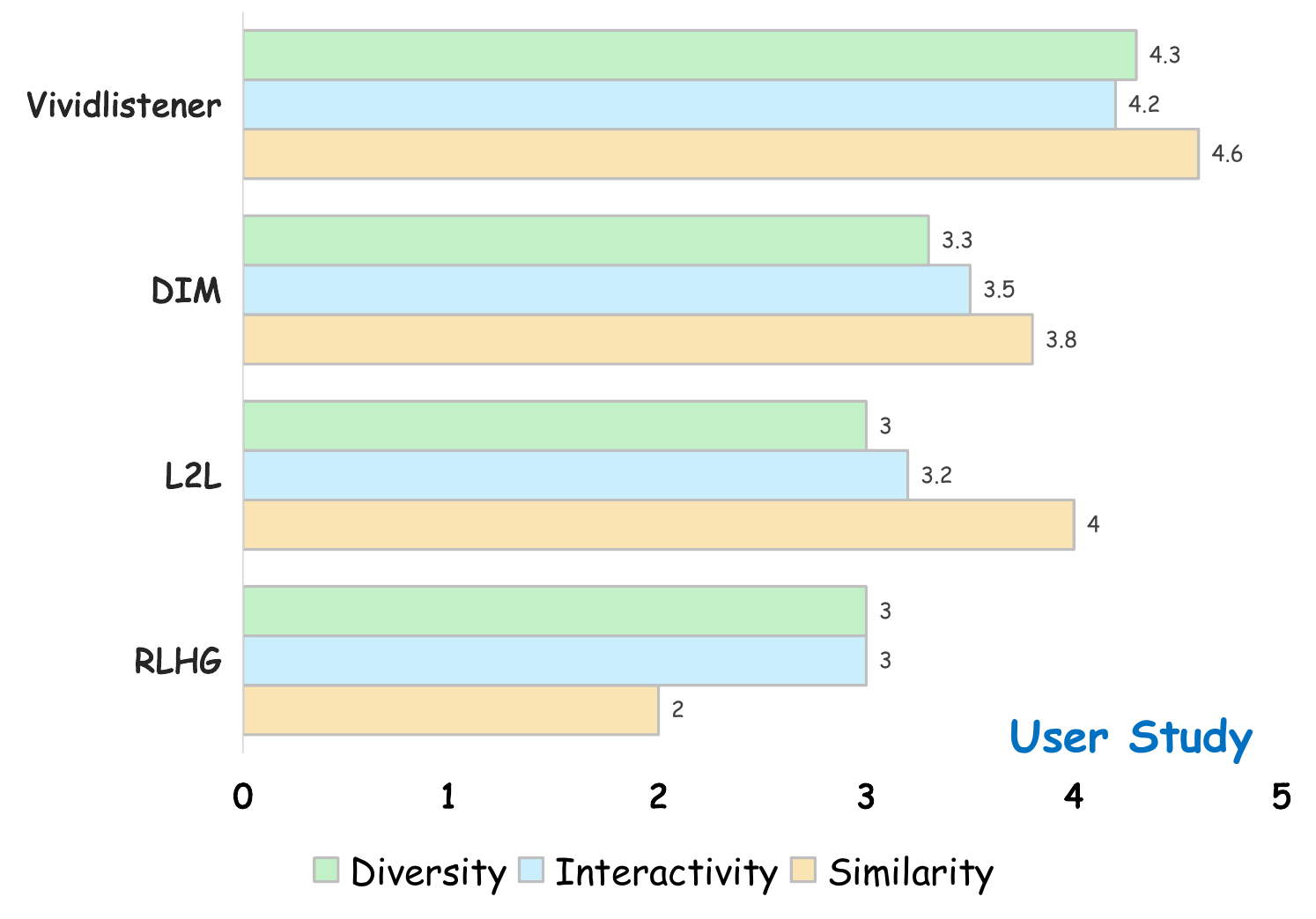}
    \caption{\textbf{User Study.} Comparisons with SOTA methods.}
    \label{fig:fig7}
    \vspace{-5mm}
\end{figure}

\noindent  \textbf{Conditional Control Results.} 
Fig.~\ref{fig:fig7} illustrates listener head motion generation results under different text-based conditions and emotional intensity controls. By incorporating open-vocabulary textual input conditions, our method surpasses the limitations of fixed-text descriptions in traditional datasets. For instance, Description 1 captures a gradual transition from a "somber expression" to a "brightened mood," while Description 2 showcases richer emotional fluctuations, demonstrating the model's adaptability and capability to generate diverse and nuanced dynamics.


\noindent  \textbf{User Study}
To further analyze the generated quality by our method and other counterparts, as well as the result alignment with given conditions, we conduct a user study involving 25 volunteers. 
The statistical results are shown in Fig.~\ref{fig:fig7}. 
All volunteers are recruited anonymously and represented diverse academic disciplines. 
Each participant is required to rate the randomly selected visualization videos from 0 (worst) to 5 (best) in terms of similarity to GT, interactivity, and diversity.
Our framework demonstrates superior performance among all competitors and receives strong user affirmation for its alignment with input conditions.

\section{Conclusion}
In this paper, we propose VividListener, a novel framework that enables expressive and controllable listener dynamics modeling for multi-modal responsive interaction.
Meanwhile, a large-scale 3D dyadic conversation dataset, ListenerX, with fine-grained multi-modal conditions, is introduced for this task.
Moreover, we propose Responsive Interaction Module for listener-speaker alignment and integration, and design Emotional Intensity Tags for emotion intensity editing.
Experimental results on ListenerX demonstrate that VividListener achieves state-of-the-art performance for expressive and controllable listener reaction generation.
In the future, we will focus on interaction tasks in more complex scenarios, such as simultaneously generating speakers and listeners in dialogue scenes, rather than responsive interaction.

\clearpage
\newpage
{
    \small
    \bibliographystyle{ieeenat_fullname}

\begin{thebibliography}{59}
\providecommand{\natexlab}[1]{#1}
\providecommand{\url}[1]{\texttt{#1}}
\expandafter\ifx\csname urlstyle\endcsname\relax
  \providecommand{\doi}[1]{doi: #1}\else
  \providecommand{\doi}{doi: \begingroup \urlstyle{rm}\Url}\fi

\bibitem[Aneja et~al.(2024)Aneja, Thies, Dai, and Nie{\ss}ner]{aneja2024facetalk}
Shivangi Aneja, Justus Thies, Angela Dai, and Matthias Nie{\ss}ner.
\newblock Facetalk: Audio-driven motion diffusion for neural parametric head models.
\newblock In \emph{Proceedings of the IEEE/CVF Conference on Computer Vision and Pattern Recognition}, pages 21263--21273, 2024.

\bibitem[Bohus and Horvitz(2010{\natexlab{a}})]{bohus2010challenges}
Dan Bohus and Eric Horvitz.
\newblock On the challenges and opportunities of physically situated dialog.
\newblock In \emph{2010 AAAI Fall Symposium Series}, 2010{\natexlab{a}}.

\bibitem[Bohus and Horvitz(2010{\natexlab{b}})]{bohus2010facilitating}
Dan Bohus and Eric Horvitz.
\newblock Facilitating multiparty dialog with gaze, gesture, and speech.
\newblock In \emph{International Conference on Multimodal Interfaces and the Workshop on Machine Learning for Multimodal Interaction}, pages 1--8, 2010{\natexlab{b}}.

\bibitem[Chang et~al.(2024)Chang, Yin, Li, Tran, and Soleymani]{chang2024libreface}
Di Chang, Yufeng Yin, Zongjian Li, Minh Tran, and Mohammad Soleymani.
\newblock Libreface: An open-source toolkit for deep facial expression analysis.
\newblock In \emph{Proceedings of the IEEE/CVF Winter Conference on Applications of Computer Vision}, pages 8205--8215, 2024.

\bibitem[Chen et~al.(2021)Chen, Tran-Thien-Y, and Florence]{chen2021usability}
Ja-Shen Chen, Le Tran-Thien-Y, and Devina Florence.
\newblock Usability and responsiveness of artificial intelligence chatbot on online customer experience in e-retailing.
\newblock \emph{International Journal of Retail \& Distribution Management}, 49\penalty0 (11):\penalty0 1512--1531, 2021.

\bibitem[Chung and Zisserman(2016)]{Chung16a}
J.~S. Chung and A. Zisserman.
\newblock Out of time: automated lip sync in the wild.
\newblock In \emph{Workshop on Multi-view Lip-reading, ACCV}, 2016.

\bibitem[Chung and Zisserman(2017)]{chung2017out}
Joon~Son Chung and Andrew Zisserman.
\newblock Out of time: automated lip sync in the wild.
\newblock In \emph{Computer Vision--ACCV 2016 Workshops: ACCV 2016 International Workshops, Taipei, Taiwan, November 20-24, 2016, Revised Selected Papers, Part II 13}, pages 251--263. Springer, 2017.

\bibitem[Cudeiro et~al.(2019)Cudeiro, Bolkart, Laidlaw, Ranjan, and Black]{cudeiro2019capture}
Daniel Cudeiro, Timo Bolkart, Cassidy Laidlaw, Anurag Ranjan, and Michael~J Black.
\newblock Capture, learning, and synthesis of 3d speaking styles.
\newblock In \emph{Proceedings of the IEEE/CVF conference on computer vision and pattern recognition}, pages 10101--10111, 2019.

\bibitem[Dan{\v{e}}{\v{c}}ek et~al.(2022)Dan{\v{e}}{\v{c}}ek, Black, and Bolkart]{danvevcek2022emoca}
Radek Dan{\v{e}}{\v{c}}ek, Michael~J Black, and Timo Bolkart.
\newblock Emoca: Emotion driven monocular face capture and animation.
\newblock In \emph{Proceedings of the IEEE/CVF Conference on Computer Vision and Pattern Recognition}, pages 20311--20322, 2022.

\bibitem[Dan{\v{e}}{\v{c}}ek et~al.(2023)Dan{\v{e}}{\v{c}}ek, Chhatre, Tripathi, Wen, Black, and Bolkart]{danvevcek2023emotional}
Radek Dan{\v{e}}{\v{c}}ek, Kiran Chhatre, Shashank Tripathi, Yandong Wen, Michael Black, and Timo Bolkart.
\newblock Emotional speech-driven animation with content-emotion disentanglement.
\newblock In \emph{SIGGRAPH Asia 2023 Conference Papers}, pages 1--13, 2023.

\bibitem[Duan et~al.(2022)Duan, Yu, Tan, Zhu, and Tan]{duan2022survey}
Jiafei Duan, Samson Yu, Hui~Li Tan, Hongyuan Zhu, and Cheston Tan.
\newblock A survey of embodied ai: From simulators to research tasks.
\newblock \emph{IEEE Transactions on Emerging Topics in Computational Intelligence}, 6\penalty0 (2):\penalty0 230--244, 2022.

\bibitem[Ekman and Friesen(1978)]{ekman1978facial}
Paul Ekman and Wallace~V Friesen.
\newblock Facial action coding system.
\newblock \emph{Environmental Psychology \& Nonverbal Behavior}, 1978.

\bibitem[Farouk(2022)]{farouk2022studying}
Maged Farouk.
\newblock Studying human robot interaction and its characteristics.
\newblock \emph{International Journal of Computations, Information and Manufacturing (IJCIM)}, 2\penalty0 (1), 2022.

\bibitem[Genay et~al.(2021)Genay, L{\'e}cuyer, and Hachet]{genay2021being}
Ad{\'e}la{\"\i}de Genay, Anatole L{\'e}cuyer, and Martin Hachet.
\newblock Being an avatar “for real”: a survey on virtual embodiment in augmented reality.
\newblock \emph{IEEE Transactions on Visualization and Computer Graphics}, 28\penalty0 (12):\penalty0 5071--5090, 2021.

\bibitem[Geng et~al.(2023)Geng, Teotia, Tendulkar, Menon, and Vondrick]{geng2023affective}
Scott Geng, Revant Teotia, Purva Tendulkar, Sachit Menon, and Carl Vondrick.
\newblock Affective faces for goal-driven dyadic communication.
\newblock \emph{arXiv preprint arXiv:2301.10939}, 2023.

\bibitem[Heusel et~al.(2017)Heusel, Ramsauer, Unterthiner, Nessler, and Hochreiter]{heusel2017gans}
Martin Heusel, Hubert Ramsauer, Thomas Unterthiner, Bernhard Nessler, and Sepp Hochreiter.
\newblock Gans trained by a two time-scale update rule converge to a local nash equilibrium.
\newblock \emph{Advances in neural information processing systems}, 30, 2017.

\bibitem[Ho et~al.(2020)Ho, Jain, and Abbeel]{ho2020denoising}
Jonathan Ho, Ajay Jain, and Pieter Abbeel.
\newblock Denoising diffusion probabilistic models.
\newblock \emph{Advances in neural information processing systems}, 33:\penalty0 6840--6851, 2020.

\bibitem[Huang and Belongie(2017)]{huang2017arbitrary}
Xun Huang and Serge Belongie.
\newblock Arbitrary style transfer in real-time with adaptive instance normalization.
\newblock In \emph{Proceedings of the IEEE international conference on computer vision}, pages 1501--1510, 2017.

\bibitem[Huang and Khan(2017)]{huang2017dyadgan}
Yuchi Huang and Saad~M Khan.
\newblock Dyadgan: Generating facial expressions in dyadic interactions.
\newblock In \emph{Proceedings of the IEEE Conference on Computer Vision and Pattern Recognition Workshops}, pages 11--18, 2017.

\bibitem[LeCun et~al.(1998)LeCun, Bottou, Bengio, and Haffner]{lecun1998gradient}
Yann LeCun, L{\'e}on Bottou, Yoshua Bengio, and Patrick Haffner.
\newblock Gradient-based learning applied to document recognition.
\newblock \emph{Proceedings of the IEEE}, 86\penalty0 (11):\penalty0 2278--2324, 1998.

\bibitem[Li et~al.(2017)Li, Bolkart, Black, Li, and Romero]{li2017learning}
Tianye Li, Timo Bolkart, Michael~J Black, Hao Li, and Javier Romero.
\newblock Learning a model of facial shape and expression from 4d scans.
\newblock \emph{ACM Trans. Graph.}, 36\penalty0 (6):\penalty0 194--1, 2017.

\bibitem[Liu et~al.(2024{\natexlab{a}})Liu, Li, Wu, and Lee]{liu2024visual}
Haotian Liu, Chunyuan Li, Qingyang Wu, and Yong~Jae Lee.
\newblock Visual instruction tuning.
\newblock \emph{Advances in neural information processing systems}, 36, 2024{\natexlab{a}}.

\bibitem[Liu et~al.(2023)Liu, Wang, Fu, Chai, Yu, Dai, and Han]{liu2023mfr}
Jin Liu, Xi Wang, Xiaomeng Fu, Yesheng Chai, Cai Yu, Jiao Dai, and Jizhong Han.
\newblock Mfr-net: Multi-faceted responsive listening head generation via denoising diffusion model.
\newblock In \emph{Proceedings of the 31st ACM International Conference on Multimedia}, pages 6734--6743, 2023.

\bibitem[Liu et~al.(2024{\natexlab{b}})Liu, Guo, Zhen, Li, Ao, and Yan]{liu2024customlistener}
Xi Liu, Ying Guo, Cheng Zhen, Tong Li, Yingying Ao, and Pengfei Yan.
\newblock Customlistener: Text-guided responsive interaction for user-friendly listening head generation.
\newblock In \emph{Proceedings of the IEEE/CVF Conference on Computer Vision and Pattern Recognition}, pages 2415--2424, 2024{\natexlab{b}}.

\bibitem[Loshchilov(2017)]{loshchilov2017decoupled}
I Loshchilov.
\newblock Decoupled weight decay regularization.
\newblock \emph{arXiv preprint arXiv:1711.05101}, 2017.

\bibitem[Messinger et~al.(2009)Messinger, Mahoor, Chow, and Cohn]{messinger2009automated}
Daniel~S Messinger, Mohammad~H Mahoor, Sy-Miin Chow, and Jeffrey~F Cohn.
\newblock Automated measurement of facial expression in infant--mother interaction: A pilot study.
\newblock \emph{Infancy}, 14\penalty0 (3):\penalty0 285--305, 2009.

\bibitem[Ng et~al.(2022)Ng, Joo, Hu, Li, Darrell, Kanazawa, and Ginosar]{ng2022learning}
Evonne Ng, Hanbyul Joo, Liwen Hu, Hao Li, Trevor Darrell, Angjoo Kanazawa, and Shiry Ginosar.
\newblock Learning to listen: Modeling non-deterministic dyadic facial motion.
\newblock In \emph{Proceedings of the IEEE/CVF Conference on Computer Vision and Pattern Recognition}, pages 20395--20405, 2022.

\bibitem[Ng et~al.(2023)Ng, Subramanian, Klein, Kanazawa, Darrell, and Ginosar]{ng2023can}
Evonne Ng, Sanjay Subramanian, Dan Klein, Angjoo Kanazawa, Trevor Darrell, and Shiry Ginosar.
\newblock Can language models learn to listen?
\newblock In \emph{Proceedings of the IEEE/CVF International Conference on Computer Vision}, pages 10083--10093, 2023.

\bibitem[Nichol and Dhariwal(2021)]{nichol2021improved}
Alexander~Quinn Nichol and Prafulla Dhariwal.
\newblock Improved denoising diffusion probabilistic models.
\newblock In \emph{International conference on machine learning}, pages 8162--8171. PMLR, 2021.

\bibitem[Nojavanasghari et~al.(2018)Nojavanasghari, Huang, and Khan]{nojavanasghari2018interactive}
Behnaz Nojavanasghari, Yuchi Huang, and Saad Khan.
\newblock Interactive generative adversarial networks for facial expression generation in dyadic interactions.
\newblock \emph{arXiv preprint arXiv:1801.09092}, 2018.

\bibitem[Peebles and Xie(2023)]{peebles2023scalable}
William Peebles and Saining Xie.
\newblock Scalable diffusion models with transformers.
\newblock In \emph{Proceedings of the IEEE/CVF International Conference on Computer Vision}, pages 4195--4205, 2023.

\bibitem[Pei et~al.(2024)Pei, Zhang, Hu, Zhang, Wang, Wu, Zhai, Yang, Shen, and Tao]{pei2024deepfake}
Gan Pei, Jiangning Zhang, Menghan Hu, Zhenyu Zhang, Chengjie Wang, Yunsheng Wu, Guangtao Zhai, Jian Yang, Chunhua Shen, and Dacheng Tao.
\newblock Deepfake generation and detection: A benchmark and survey.
\newblock \emph{arXiv preprint arXiv:2403.17881}, 2024.

\bibitem[Peng et~al.(2023)Peng, Wu, Song, Xu, Zhu, He, Liu, and Fan]{peng2023emotalk}
Ziqiao Peng, Haoyu Wu, Zhenbo Song, Hao Xu, Xiangyu Zhu, Jun He, Hongyan Liu, and Zhaoxin Fan.
\newblock Emotalk: Speech-driven emotional disentanglement for 3d face animation.
\newblock In \emph{Proceedings of the IEEE/CVF International Conference on Computer Vision}, pages 20687--20697, 2023.

\bibitem[Qi et~al.(2024{\natexlab{a}})Qi, Liu, Li, Hou, Xin, and Yu]{qi2024emotiongesture}
Xingqun Qi, Chen Liu, Lincheng Li, Jie Hou, Haoran Xin, and Xin Yu.
\newblock Emotiongesture: Audio-driven diverse emotional co-speech 3d gesture generation.
\newblock \emph{IEEE Transactions on Multimedia}, 2024{\natexlab{a}}.

\bibitem[Qi et~al.(2024{\natexlab{b}})Qi, Pan, Li, Yuan, Chi, Li, Luo, Xue, Zhang, Liu, et~al.]{qi2024weakly}
Xingqun Qi, Jiahao Pan, Peng Li, Ruibin Yuan, Xiaowei Chi, Mengfei Li, Wenhan Luo, Wei Xue, Shanghang Zhang, Qifeng Liu, et~al.
\newblock Weakly-supervised emotion transition learning for diverse 3d co-speech gesture generation.
\newblock In \emph{Proceedings of the IEEE/CVF Conference on Computer Vision and Pattern Recognition}, pages 10424--10434, 2024{\natexlab{b}}.

\bibitem[Qi et~al.(2024{\natexlab{c}})Qi, Zhang, Wang, Pan, Liu, Li, Chi, Li, Zhang, Xue, et~al.]{qi2024cocogesture}
Xingqun Qi, Hengyuan Zhang, Yatian Wang, Jiahao Pan, Chen Liu, Peng Li, Xiaowei Chi, Mengfei Li, Qixun Zhang, Wei Xue, et~al.
\newblock Cocogesture: Toward coherent co-speech 3d gesture generation in the wild.
\newblock \emph{arXiv preprint arXiv:2405.16874}, 2024{\natexlab{c}}.

\bibitem[Radford et~al.(2021)Radford, Kim, Hallacy, Ramesh, Goh, Agarwal, Sastry, Askell, Mishkin, Clark, et~al.]{radford2021learning}
Alec Radford, Jong~Wook Kim, Chris Hallacy, Aditya Ramesh, Gabriel Goh, Sandhini Agarwal, Girish Sastry, Amanda Askell, Pamela Mishkin, Jack Clark, et~al.
\newblock Learning transferable visual models from natural language supervision.
\newblock In \emph{International conference on machine learning}, pages 8748--8763. PMLR, 2021.

\bibitem[Richard et~al.(2021)Richard, Zollh{\"o}fer, Wen, De~la Torre, and Sheikh]{richard2021meshtalk}
Alexander Richard, Michael Zollh{\"o}fer, Yandong Wen, Fernando De~la Torre, and Yaser Sheikh.
\newblock Meshtalk: 3d face animation from speech using cross-modality disentanglement.
\newblock In \emph{Proceedings of the IEEE/CVF International Conference on Computer Vision}, pages 1173--1182, 2021.

\bibitem[Riehle et~al.(2017)Riehle, Kempkensteffen, and Lincoln]{riehle2017quantifying}
Marcel Riehle, J{\"u}rgen Kempkensteffen, and Tania~M Lincoln.
\newblock Quantifying facial expression synchrony in face-to-face dyadic interactions: Temporal dynamics of simultaneously recorded facial emg signals.
\newblock \emph{Journal of Nonverbal Behavior}, 41:\penalty0 85--102, 2017.

\bibitem[Russell(1980)]{russell1980circumplex}
James~A Russell.
\newblock A circumplex model of affect.
\newblock \emph{Journal of personality and social psychology}, 39\penalty0 (6):\penalty0 1161, 1980.

\bibitem[Song et~al.(2020)Song, Meng, and Ermon]{song2020denoising}
Jiaming Song, Chenlin Meng, and Stefano Ermon.
\newblock Denoising diffusion implicit models.
\newblock \emph{arXiv preprint arXiv:2010.02502}, 2020.

\bibitem[Song et~al.(2023)Song, Yin, Jin, Dong, and Xu]{song2023emotional}
Luchuan Song, Guojun Yin, Zhenchao Jin, Xiaoyi Dong, and Chenliang Xu.
\newblock Emotional listener portrait: Neural listener head generation with emotion.
\newblock In \emph{Proceedings of the IEEE/CVF International Conference on Computer Vision}, pages 20839--20849, 2023.

\bibitem[Sonlu et~al.(2021)Sonlu, G{\"u}d{\"u}kbay, and Durupinar]{sonlu2021conversational}
Sinan Sonlu, U{\u{g}}ur G{\"u}d{\"u}kbay, and Funda Durupinar.
\newblock A conversational agent framework with multi-modal personality expression.
\newblock \emph{ACM Transactions on Graphics (TOG)}, 40\penalty0 (1):\penalty0 1--16, 2021.

\bibitem[Soori et~al.(2023)Soori, Arezoo, and Dastres]{soori2023artificial}
Mohsen Soori, Behrooz Arezoo, and Roza Dastres.
\newblock Artificial intelligence, machine learning and deep learning in advanced robotics, a review.
\newblock \emph{Cognitive Robotics}, 3:\penalty0 54--70, 2023.

\bibitem[Stergiou and Poppe(2019)]{stergiou2019analyzing}
Alexandros Stergiou and Ronald Poppe.
\newblock Analyzing human--human interactions: A survey.
\newblock \emph{Computer Vision and Image Understanding}, 188:\penalty0 102799, 2019.

\bibitem[Sun et~al.(2024{\natexlab{a}})Sun, Chu, Zhou, Wang, and Koike]{sun2024avi}
Yasheng Sun, Wenqing Chu, Hang Zhou, Kaisiyuan Wang, and Hideki Koike.
\newblock Avi-talking: Learning audio-visual instructions for expressive 3d talking face generation.
\newblock \emph{IEEE Access}, 2024{\natexlab{a}}.

\bibitem[Sun et~al.(2024{\natexlab{b}})Sun, Lv, Ye, Lin, Sheng, Wen, Yu, and Liu]{sun2024diffposetalk}
Zhiyao Sun, Tian Lv, Sheng Ye, Matthieu Lin, Jenny Sheng, Yu-Hui Wen, Minjing Yu, and Yong-jin Liu.
\newblock Diffposetalk: Speech-driven stylistic 3d facial animation and head pose generation via diffusion models.
\newblock \emph{ACM Transactions on Graphics (TOG)}, 43\penalty0 (4):\penalty0 1--9, 2024{\natexlab{b}}.

\bibitem[Tamon et~al.(2024)Tamon, Yasuhisa, Yukoh, Kengo, Ryota, and Norihide]{tamon2024listening}
Mikawa Tamon, Fujii Yasuhisa, Wakabayashi Yukoh, Ohta Kengo, Nishimura Ryota, and Kitaoka Norihide.
\newblock Listening head motion generation for multimodal dialog system.
\newblock In \emph{2024 11th International Conference on Advanced Informatics: Concept, Theory and Application (ICAI)}, pages 1--6. IEEE, 2024.

\bibitem[Thambiraja et~al.(2023)Thambiraja, Habibie, Aliakbarian, Cosker, Theobalt, and Thies]{thambiraja2023imitator}
Balamurugan Thambiraja, Ikhsanul Habibie, Sadegh Aliakbarian, Darren Cosker, Christian Theobalt, and Justus Thies.
\newblock Imitator: Personalized speech-driven 3d facial animation.
\newblock In \emph{Proceedings of the IEEE/CVF International Conference on Computer Vision}, pages 20621--20631, 2023.

\bibitem[Toisoul et~al.(2021)Toisoul, Kossaifi, Bulat, Tzimiropoulos, and Pantic]{toisoul2021estimation}
Antoine Toisoul, Jean Kossaifi, Adrian Bulat, Georgios Tzimiropoulos, and Maja Pantic.
\newblock Estimation of continuous valence and arousal levels from faces in naturalistic conditions.
\newblock \emph{Nature Machine Intelligence}, 3\penalty0 (1):\penalty0 42--50, 2021.

\bibitem[Tran et~al.(2024{\natexlab{a}})Tran, Chang, Siniukov, and Soleymani]{tran2024dim}
Minh Tran, Di Chang, Maksim Siniukov, and Mohammad Soleymani.
\newblock Dim: Dyadic interaction modeling for social behavior generation.
\newblock In \emph{European Conference on Computer Vision}, pages 484--503. Springer, 2024{\natexlab{a}}.

\bibitem[Tran et~al.(2024{\natexlab{b}})Tran, Chang, Siniukov, and Soleymani]{tran2024dyadic}
Minh Tran, Di Chang, Maksim Siniukov, and Mohammad Soleymani.
\newblock Dyadic interaction modeling for social behavior generation.
\newblock \emph{arXiv preprint arXiv:2403.09069}, 2024{\natexlab{b}}.

\bibitem[Van Den~Oord et~al.(2017)Van Den~Oord, Vinyals, et~al.]{van2017neural}
Aaron Van Den~Oord, Oriol Vinyals, et~al.
\newblock Neural discrete representation learning.
\newblock \emph{Advances in neural information processing systems}, 30, 2017.

\bibitem[Volonte et~al.(2020)Volonte, Hsu, Liu, Mazer, Wong, and Babu]{volonte2020effects}
Matias Volonte, Yu-Chun Hsu, Kuan-Yu Liu, Joe~P Mazer, Sai-Keung Wong, and Sabarish~V Babu.
\newblock Effects of interacting with a crowd of emotional virtual humans on users’ affective and non-verbal behaviors.
\newblock In \emph{2020 IEEE Conference on Virtual Reality and 3D User Interfaces (VR)}, pages 293--302. IEEE, 2020.

\bibitem[Wang et~al.(2023)Wang, Wu, Xing, and Jia]{wang2023versatile}
Haoyu Wang, Haozhe Wu, Junliang Xing, and Jia Jia.
\newblock Versatile face animator: Driving arbitrary 3d facial avatar in rgbd space.
\newblock In \emph{Proceedings of the 31st ACM International Conference on Multimedia}, pages 7776--7784, 2023.

\bibitem[Wu et~al.(2023)Wu, Zhou, Jia, Xing, Wen, and Wen]{wu2023speech}
Haozhe Wu, Songtao Zhou, Jia Jia, Junliang Xing, Qi Wen, and Xiang Wen.
\newblock Speech-driven 3d face animation with composite and regional facial movements.
\newblock In \emph{Proceedings of the 31st ACM International Conference on Multimedia}, pages 6822--6830, 2023.

\bibitem[Zhang et~al.(2022)Zhang, Cai, Pan, Hong, Guo, Yang, and Liu]{zhang2022motiondiffuse}
Mingyuan Zhang, Zhongang Cai, Liang Pan, Fangzhou Hong, Xinying Guo, Lei Yang, and Ziwei Liu.
\newblock Motiondiffuse: Text-driven human motion generation with diffusion model.
\newblock \emph{arXiv preprint arXiv:2208.15001}, 2022.

\bibitem[Zhao et~al.(2024)Zhao, Long, Zhang, Qin, Liang, Zhang, Zhang, Yu, and Xu]{zhao2024media2face}
Qingcheng Zhao, Pengyu Long, Qixuan Zhang, Dafei Qin, Han Liang, Longwen Zhang, Yingliang Zhang, Jingyi Yu, and Lan Xu.
\newblock Media2face: Co-speech facial animation generation with multi-modality guidance.
\newblock In \emph{ACM SIGGRAPH 2024 conference papers}, pages 1--13, 2024.

\bibitem[Zhou et~al.(2022)Zhou, Bai, Zhang, Yao, Zhao, and Mei]{zhou2022responsive}
Mohan Zhou, Yalong Bai, Wei Zhang, Ting Yao, Tiejun Zhao, and Tao Mei.
\newblock Responsive listening head generation: a benchmark dataset and baseline.
\newblock In \emph{European Conference on Computer Vision}, pages 124--142. Springer, 2022.

\end{thebibliography}

}

\newpage
\section{Supplementary material}
\subsection{Implementation Details}
In our experiments, we set the total generated sequence length N = 240 with the normalized fps=30. 
Thus, the duration of each sequence is 8 seconds. 
\(S_a\) is initially represented as audio signal waves, which are converted into mel-spectrograms with an FFT window size of 512, and a hop length=160. 
The dimension of input audio mel-spectrograms is 128 × 480. 
The textual descriptions $text$ of the listener are encoded with the CLIP model~\cite{radford2021learning}. 
The listener emotional intensity tags \( \sigma \) are obtained at the frequency of 5 Hz.

In the training stage, we empirically set \(\lambda_{\text{simple}}\)=2, \(\lambda_{\text{emotional}}\)=0.2, \(\lambda_{\text{vel}}\)=0.8. The initial learning rate is set to \(1 \times 10^{-5}\) with AdamW optimizer~\cite{loshchilov2017decoupled}. 
Similar to DDPM~\cite{nichol2021improved}, we set the diffusion timestep as 1,000 with the cosine noisy schedule. Our model is trained on 8 NVIDIA V100 GPUs with a total batch size of 64. The training process spans 800 epochs and is completed in 1 days. 
During inference, we adopt DDIM ~\cite{song2020denoising} sampling strategy with 100 denoising timesteps to produce head movements. 
Our VividListener model synthesizes head movements incorporating 50-dimensional expression parameters and 6-dimensional pose features. 
Thus, the generated sequence has a dimension of \( \mathbb{R}^{240 \times 56} \), where 240 denotes the frames and 56 means the head representation.

\subsection{Metrics details}
To comprehensively evaluate the realism, synchrony, and diversity of the generated listener dynamics, we introduce several metrics.
\begin{itemize}[leftmargin=*]
 \item  \textbf{Fréchet Distance(FD) for Motion Realism}. To quantify the distribution discrepancy between the generated and ground-truth motions ~\cite{heusel2017gans}.
\item   \textbf{Paired FD(P-FD) for Synchrony}. To quantify the quality and synchrony of the listener and speaker dynamics based on the distribution distances of listener-speaker pairs~\cite{heusel2017gans}.

\item  \textbf{Mean Squared Error (MSE) for Accuracy}. To quantify the accuracy between generated and ground-truth motions.

\item \textbf{Shannon Index (SID) for Diversity}. To quantify diversity of motion predictions with k-means clustering, where the histogram entropy of cluster IDs within each sequence is calculated.

\item \textbf{Variance (Var)}: Following L2L~\cite{ng2022learning}, we use this metric to measure the variance of the listener motion in time series.

\item \textbf{Residual Pearson Correlation Coefficient (rPCC)}: rPCC evaluates the frame-by-frame correlation between listener and speaker movements~\cite{messinger2009automated, riehle2017quantifying}. 

\end{itemize}

\subsection{Qualitative analysis of cross-scenario}
\begin{figure*}[t]
    \centering
    \includegraphics[width=0.7\textwidth]{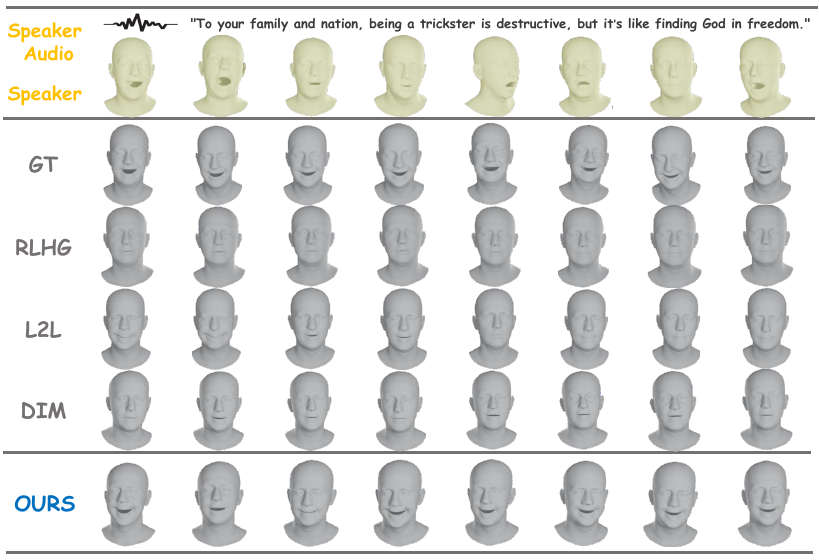}
    \caption{\textbf{Visual comparisons on cross-scenario.} We present visualizations of listener motions generated in a cross-scenario setting, where all models are trained on the DailyX subset and tested on the InterviewX dataset.}
    \label{fig:fig5}
\end{figure*}

In cross-scenario inference,as shown in Fig.~\ref{fig:fig5}, the results produced by the aforementioned methods exhibit significant discrepancies compared to the Ground Truth. 
This may be attributed to their focus on single-scene or short-sequence data, which hinders their ability to model the complex characteristics of long-sequence motions. 
In contrast, our method effectively models multi-modal conditions and captures global semantics in long sequences, enabling the long-term generation of vivid and dynamic listener head motions. 
This further highlights the generalization capability to finely capture emotional dynamics over extended sequences.

\subsection{Multi-turn Interaction}
\begin{figure*}[t]
    \centering
    \includegraphics[width=0.72\textwidth]{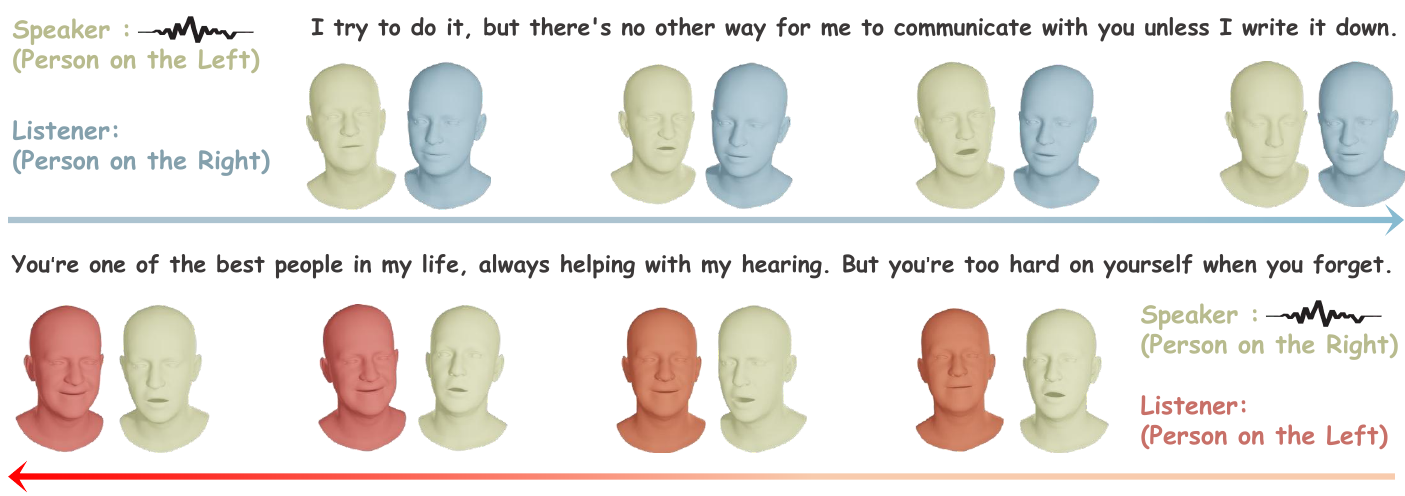}
    \caption{\textbf{Multi-turn Interaction Result.} We generated the results of a dyadic multi-turn conversation, where the roles of the speaker and listener dynamically alternate. (In the first row) the person on the left speaks, while the person on the right listens. (In the second row) the roles reverse, with the person on the right speaking and the person on the left responding naturally.}
    \label{fig:fig6}
\end{figure*}

To demonstrate the interactivity of our method and its applicability in real-world scenarios, we conduct our model on multi-turn interaction scenes.
As shown in Fig.~\ref{fig:fig6}, our framework effectively captures the temporal dependencies and emotional shifts in long-term multi-turn conversations, ensuring smooth and natural transitions between listener reactions. For more visualization demo videos please refer to our supplementary material.

\end{document}